\documentclass[10pt,twocolumn,letterpaper]{article}

\usepackage{wacv}
\usepackage[accsupp]{axessibility}

\usepackage{graphicx}
\usepackage{amsmath}
\usepackage{amsfonts}
\usepackage{booktabs}

\usepackage{multirow}

\usepackage{xcolor}
\usepackage{bbm}
\definecolor{applegreen}{rgb}{0.0, 0.5, 0.0}

\usepackage[pagebackref,breaklinks,colorlinks]{hyperref}

\usepackage[capitalize]{cleveref}
\crefname{section}{Sec.}{Secs.}
\Crefname{section}{Section}{Sections}
\Crefname{table}{Table}{
Tables}
\crefname{table}{Tab.}{Tabs.}

\def\wacvPaperID{1570} 
\def\confName{WACV}
\def\confYear{2025}

\begin{document}

\title{MixDiff: Mixing Natural and Synthetic Images for \\Robust Self-Supervised Representations}

\author{Reza Akbarian Bafghi$^{1}$\thanks{Joint first-authorship.}
\and
Nidhin Harilal$^{1}$\footnotemark[1]
\and
Claire Monteleoni$^{1,2}$
\and
Maziar Raissi$^{
3}$
\and
$^{1}$ University of Colorado, Boulder
\and
$^{2}$ INRIA, Paris
\and
$^{3}$ University of California, Riverside
\and
{\tt\small
\{reza.akbarianbafghi,
nidhin.harilal,
cmontel\}@colorado.edu, maziar.raissi@ucr.edu
}
}
\maketitle

\begin{abstract}
This paper introduces \textit{MixDiff}, a new self-supervised learning (SSL) pre-training framework that combines real and synthetic images. Unlike traditional SSL methods that predominantly use real images, MixDiff uses a variant of Stable Diffusion to replace an augmented instance of a real image, facilitating the learning of cross real-synthetic image representations.  Our key insight is that while models trained solely on synthetic images underperform, combining real and synthetic data leads to more robust and adaptable representations. Experiments show MixDiff enhances SimCLR, BarlowTwins, and DINO across various robustness datasets and domain transfer tasks, boosting SimCLR's ImageNet-1K accuracy by 4.56\%. Our framework also demonstrates comparable performance without needing any augmentations, a surprising finding in SSL where augmentations are typically crucial. Furthermore, MixDiff achieves similar results to SimCLR while requiring less real data, highlighting its efficiency in representation learning\footnote{We have made the source code and generated data available to the public at: \url{https://github.com/cryptonymous9/mixing-ssl}}.

\end{abstract}


\vspace{-5pt}
\section{Introduction}

\begin{figure}
     \centering
     \includegraphics[width=\linewidth]{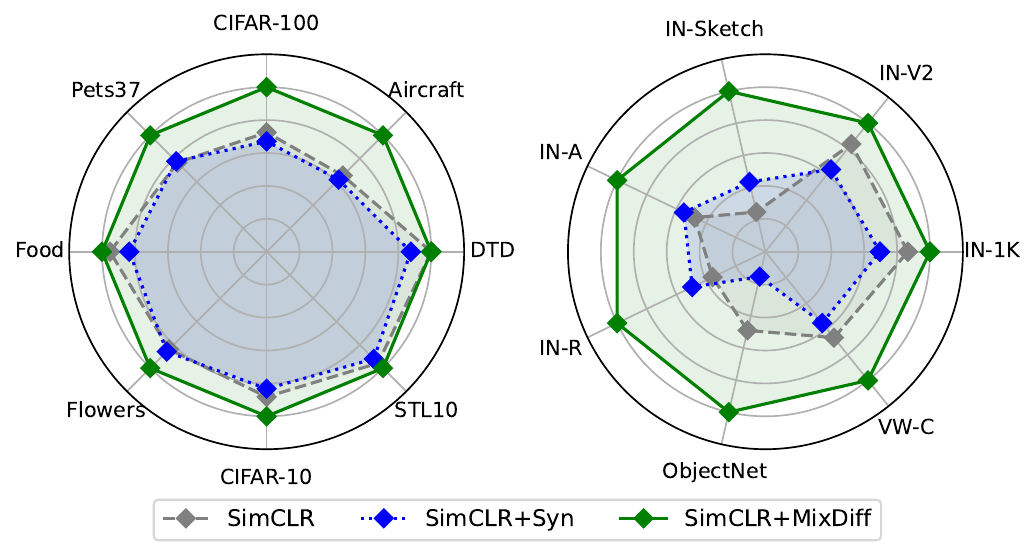}
    \vspace{-0.5cm}
    \caption{Comparison of SimCLR performance on real, synthetic (Syn), and mixed real and synthetic images (MixDiff). The radar charts show normalized accuracy across 8 transfer learning datasets (left) and ImageNet-1K plus 6 distribution shift datasets (right), with values from 0.5 to 1.1. MixDiff enhances in-distribution and robustness performance and generalizes better. More details in Sec.~\ref{sec:exp}.}
        \vspace{-0.4cm}

     \label{fig:intro}
\end{figure}


Self-supervised learning (SSL) has enjoyed significant advancements in recent years~\cite{gui2024survey}. The capability of joint-embedding SSL architectures to generate high-quality features using pretext tasks now parallels, and in some cases surpasses, that of supervised learning. Joint-embedding SSL methods can include distillation~\cite{xu2020knowledge, dino, byol} or contrastive strategies~\cite{le2020contrastive, simclr, moco, barlow}, where multiple network branches aim to learn representations by maximizing agreement between differently augmented views of the same data example in the embedding space. Even though such recent advancements in SSL save annotation costs, preparing training data is still challenging. Current research within SSL predominantly concentrates on the development of the pretext tasks, while the characteristics of the data being utilized for learning remain less explored. 

This focus on pretext tasks, however, has led researchers to consider alternative data sources to further improve model performance. In particular, there has been a growing interest in using synthetic images for self-supervision~\cite{wu2023synthetic, Tian2023StableRepSI, Wang2024DoGD}. The appeal of synthetic image datasets lies in their ease of generation, a wider range of semantic content, and minimal human intervention, addressing key concerns in computer vision like cost efficiency and fairness in data collection and annotation. Although generative models address the issue of data scarcity, exclusive reliance on synthetic images for supervision is not without drawbacks. The main challenge has been the domain gap between synthetic and real-world data. Models trained only on synthetic images often struggle to adapt to real-world settings due to their limited exposure to the variability and complexity of natural images~\cite{tremblay2018training}. This issue is particularly pronounced in large-scale image recognition tasks, where models trained on synthetic data typically underperform those trained on real images~\cite{ vanherle2022analysis, jeelani2023expanding}. In response to this, our research proposes a novel training framework that integrates both real and synthetic images, aiming to harness the strengths of each data source while mitigating their individual weaknesses.


  

In this paper, we introduce MixDiff to explore the potential of combining synthetic images generated by generative models without any labeled data with real-world images for SSL training. MixDiff is a simple framework that replaces an augmented instance of a real image in an existing joint-embedding SSL pipeline with a synthetic image from a generative model. The simplicity of the framework allows it to incorporate it in existing SSL methods like SimCLR~\cite{simclr}, BarlowTwins~\cite{barlow} and DINO~\cite{dino} as shown in Figure~\ref
{fig:MixDiff}. 

MixDiff works on the idea that synthetically generated images generate harder positive pairs that makes the overall training objective less trivial~\cite{robinson2021can}. 
We find that SSL models pre-trained exclusively on synthetic images underperform compared to those pre-trained with real images across most scenarios. Interestingly, our proposed MixDiff, which uses both real and synthetic images, improves model performance in SSL not only on in-distribution datasets but also on various out-of-distribution tasks as we show in Figure~\ref{fig:intro}, suggesting enhanced representation learning. Specifically, we observe an average increase in top-1 accuracy of about 26.92\% across six distributional datasets and a 7.36\% improvement in transfer learning across eight datasets. This observation suggests that while synthetic images alone may be insufficient for optimal pre-training, MixDiff capitalizes on the synergistic approach of leveraging the strengths of both image types to learn more robust SSL representations.

Building on these results, we investigated how synthetic image quality affects model performance in SSL. Prior research has shown a strong correlation between the quality of diffusion-generated synthetic images and model performance. Our study reveals that MixDiff is less sensitive to variations in synthetic image quality, potentially reducing the need for precise quality optimization. We also find that integrating synthetic data in models like DINO may de-emphasize background features, suggesting enhanced scene layout understanding beneficial for image segmentation. Despite the computational cost of generating synthetic images, MixDiff requires fewer real images to match SimCLR performance, indicating more efficient SSL pre-training. The reduced reliance on large real datasets and high-quality synthetic images, coupled with robustness to distributional shifts and transfer learning, positions MixDiff as a promising approach for enhancing SSL pre-training.
\section{Related Work}
\paragraph{Self-supervised Learning.}
While SSL techniques exist in different forms, one of the most successful self-supervised learning paradigms is joint-embedding SSL~\cite{misra2020self, moco, ye2019unsupervised, oord2018representation, simclr,barlow}. The main focus of joint-embedding SSL is instance-based discriminative learning~\cite{dosovitskiy2014discriminative, assran2023self}, where each image is considered to be its own class, and a model is trained by discriminating different views of the same image generated using data augmentation~\cite{moco,simclr,barlow,dino}. One such example is SimCLR~\cite{simclr}, which uses an InfoNCE-based formulation~\cite{oord2018representation} to bring in the representation of different views of the same image closer (positive pairs) and repel representations of views from different images (negative pairs) apart. In most cases, joint-embedding SSL methods work in a Siamese setting\cite{chen2021exploring} where two branches have identical architectures and share weights. However, networks such as the Siamese setting are vulnerable to collapsing to trivial representations. BarlowTwins~\cite{barlow} brings covariance regularization to the contrastive setting to enforce a non-collapsing solution. More recently, works such as DINO~\cite{dino} have shown alternative ways to prevent collapse using architectural strategies inspired by knowledge distillation~\cite{hinton2015distilling} and addressing catastrophic forgetting~\cite{bafghi2024parameter}. We provide an improved pre-training mechanism for representation learning in such joint-embedding SSL techniques.
\vspace{-0.4cm}
\paragraph{Learning using Synthetic Data.}
Recent advancements in machine learning have increasingly leveraged synthetic data across a variety of domains \cite{Silver2017MasteringTG,Mimura2018LeveragingSS, Meng2022GeneratingTD,hammoud2024synthclip}. This type of data is particularly crucial for tasks that demand extensive labeled datasets, such as human pose estimation \cite{Ma2022PretrainedDM, Guo2022LearningVR}, semantic segmentation \cite{Chen2018LearningSS, Rewatbowornwong2021RepurposingGF}, optical flow estimation~\cite{Kim2022HowTA,Sun2021AutoFlowLA}, and language models~\cite{abdin2024phi,fan2024scaling,Veselovsky2023GeneratingFS}. In the task of image classification, several studies have demonstrated the effectiveness of synthetic data~\cite{Wang2024EnhanceIC}. \cite{He2022IsSD,trabucco2023effective} illustrates its application in data-scarce settings and transfer learning; \cite{Wang2023BetterDM} explores its role in enhancing adversarial training; \cite{Dunlap2023DiversifyYV} diversifies images; \cite{Bansal2023LeavingRT,Sariyildiz2022FakeIT} evaluates model robustness against natural distribution shifts using synthetic data; and \cite{trabucco2023effective,Azizi2023SyntheticDF} discusses augmentation of images through fine-tuned diffusion models. It is important to note that all of these studies focus on supervised learning. Our work, however, is distinct in its concentration on SSL. Recently, there has been significant interest in leveraging synthetic data for SSL~\cite{wu2023synthetic,Tian2023StableRepSI,Wang2024DoGD}. \cite{Tian2023StableRepSI} employ text-to-image diffusion models to generate multiple images from a single caption, while our approach uses image-to-image diffusion models to produce one image per source, reducing the dependency on labeled data and associated costs. \cite{wu2023synthetic} introduces a data generation framework to enhance contrastive learning, with the generator trained with the SSL model. \cite{Wang2024DoGD} train diffusion models and mix real and synthetic data to boost contrastive learning through data augmentation and inflation. Unlike these methods, our approach does not require training a new generative model. Instead, we utilize off-the-shelf variants of Stable Diffusion\cite{Rombach2021HighResolutionIS,Xu2022VersatileDT} to improve the quality of representations in existing SSL models. 
\vspace{-0.3cm}
\paragraph{Generative Models.} The landscape of synthetic image generation has seen a significant evolution, with Generative Adversarial Networks (GANs) such as BigGAN \cite{Brock2018LargeSG} initially setting a high standard. These models have been pivotal in pushing the boundaries of image realism and quality. Recently, diffusion models have emerged as a promising alternative, demonstrating impressive results in both conditional \cite{Dhariwal2021DiffusionMB,Saharia2021ImageSV} and unconditional \cite{Ho2021CascadedDM} synthetic image generation. Text-to-image diffusion models like DALL-E \cite{Ramesh2021ZeroShotTG} and Imagen \cite{Saharia2022PhotorealisticTD} are notable examples, showcasing the ability to create detailed and contextually accurate images from textual descriptions. Our research takes a unique turn by focusing on the image-to-image diffusion model, specifically a fine-tuned version of Stable Diffusion \cite{Rombach2021HighResolutionIS}. This model distinguishes itself by utilizing CLIP \cite{Radford2021LearningTV} image embeddings instead of text embeddings.

\section{Method}\label{sec:method}
In this work, we focus on Self-Supervised Learning (SSL) techniques, particularly those that consolidate representations from different perspectives or augmentations of the same instance~\cite{alexey2016discriminative, wu2018unsupervised, moco, simclr}. The main idea behind this technique is, through iterative processes, these representations gradually become less sensitive to the transformations generating these varied views. Consequently, this leads to the learning of image representations that are notably effective for vision tasks such as classification~\cite{moco, simclr, barlow}. In this section, we introduce our framework, MixDiff, which uniquely employs both real and synthetically generated data through stable diffusion, and see how it can be incorporated in some of the existing SSL frameworks.
\subsection{Description of MixDiff}
Consider $x_1$ and $x'_1$, two augmented patches from an image, randomly selected from a dataset. These augmentations can include a variety of changes, such as altering spatial positions within an image, adding varying noise and applying random color adjustments, etc. Existing instance-based discriminative SSL methods primarily rely on real images~\cite{moco,simclr,barlow,dino}. In these methods, the representation derived from the first augmentation, $x_1$, of a real image is anticipated to closely align with the representation of the second augmentation, $x_1'$, of the same image as shown in Figure~\ref{fig:intro}. Our MixDiff framework modifies this approach by incorporating synthetically generated images alongside real ones. The primary objective of MixDiff is to synchronize the representations of real and synthetic images, thus enhancing existing SSL methodologies such as SimCLR, DINO, and BarlowTwins.

The synthetic images employed by MixDiff have a unique characteristic that they share a variation of the same semantic component or object with that of the real image. To achieve this, we employ Image Variation Diffuser ~\cite{ivd}\footnote{https://huggingface.co/lambdalabs/sd-image-variations-diffusers}, a variant of Stable Diffusion (SD)~\cite{Rombach2021HighResolutionIS} tailored to generate diverse images while preserving semantic categories or in simple terms, the image class. In the SD-based generative model, represented as $g^{k}_{SD}(\cdot)$, where $k$ indicates the guidance scale influencing the generative features from the input image, an input $x_i \sim D$ yields a synthetic counterpart $\tilde{x_i}$, such that $\tilde{x_i} = g^k_{SD}(x_i)$
The innovative aspect of MixDiff lies in substituting a portion of the augmentation process, i.e, the second branch of augmentation $x_i'$ within the SSL framework with these synthetic images $\tilde{x_i}$, to learn cross real-synthetic image representations a shown in Figure~\ref{fig:intro}.
\begin{figure*}
    \centering
    \includegraphics[width=\linewidth]{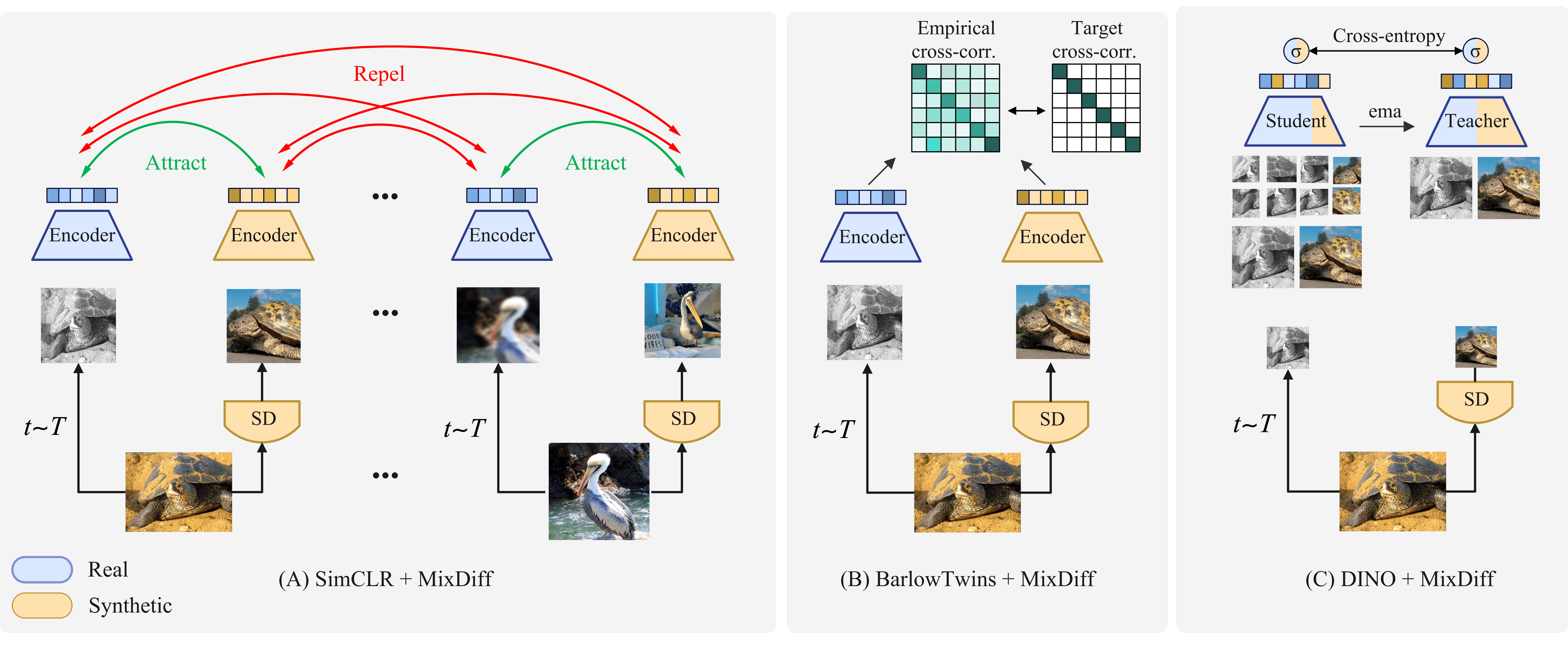}
    \vspace{-0.6cm}
    \caption{Existing SSL methods, including (A) SimCLR, (B) Barlow Twins, and (C) DINO, have been enhanced with our novel MixDiff approach. In both (A) SimCLR and (B) Barlow Twins, we replace a branch representing the positive pair with a synthetic image generated without the label using Stable Diffusion. This modification enables the learning of real-synthetic view prediction. (C) DINO utilizes a distillation framework with two global views for the teacher and a mix of two global and eight local views for the student. Our adaptation integrates a blend of global and local synthetic and real images facilitating learning correspondences between global-to-local on top of real-to-synthetic image views.}
    \label{fig:MixDiff}
         \vspace{-0.4cm}
\end{figure*}
\subsection{Mixing in joint-embedding SSL}

\paragraph{SimCLR + MixDiff:} In SimCLR's contrastive setup~\cite{simclr}, 'positive` and 'negative` pairs of images are identified, with the goal of either converging or diverging their representations. In SimCLR, two augmented views are generated for each image in a mini-batch, resulting in $2N$ images for a mini-batch size of $N$. Each view is paired with its corresponding alternate view as a 'positive` pair, while the remaining $2(N-1)$ images are treated as 'negative' pairs. We now propose to incorporate \textit{mixing} into SimCLR. We first define a new set $\{x_k , \tilde{x}_k\}$ for $k \in [1, 2, ..., N]$, where $\tilde{x}_k$ denotes the synthetically generated counterpart of $x_k$, thus establishing pairs like $x_1$ and $\tilde{x}_1$ as positive examples. The contrastive prediction task of the modified, which we term as SimCLR+MixDiff now involves identifying $x_1$ and $\tilde{x}_1$   in $\{x_k, \tilde{x}_k\} \,\forall \,k \in [1,N]$. The modified loss function for the mixed version of SimCLR, denoted as $\mathcal{L}_{MixSR}$, for a positive pair of examples ($x_i$, $\tilde{x_i}$), is defined as:
\[\mathcal{L}_{MixSR}^i \triangleq -\log\frac{\exp(\text{sim}(z_i, \tilde{z_i}))/\tau}{\Sigma_{z\in\{z_j, \tilde{z_j} \forall j\in [1,N]\}} \mathbbm{1}_{z\ne {z_i, \tilde{z_i}}} \exp(\text{sim}(z_i, z))}\]
where for a given two feature vectors, $u$ and $v$, their `sim' refers to the cosine similarity and is calculated as $\text{sim}(u, v) = \frac{u^Tv}{||u||,||v||}$, representing the dot product of the $l_2$ normalized vectors $u$ and $v$. And $z_i$ and $\tilde{z_i}$ are the outputs of the network $f(\cdot)$ we are developing, expressed as:
\[z_i = f(x_i) \;\;\text{and}\;\;\tilde{z_i} = f(\tilde{x_i})=f(g^k_{SD}(x_i))\]

\paragraph{Barlow Twins + MixDiff:} The Barlow Twins framework~\cite{barlow}, while maintaining a Siamese network structure similar to SimCLR~\cite{simclr}, adopts a distinct approach to representation alignment. The difference lies in the Barlow Twins' objective function, which assesses the cross-correlation matrix between the embeddings from two identical networks. These networks process distorted versions of a batch of samples, with the aim of aligning this matrix closely with the identity matrix. This alignment ensures that the embeddings of distorted versions of a sample are similar, while simultaneously reducing redundancy among the components of these embeddings.

In our modified approach, as we show in Figure~\ref{fig:MixDiff} (B), we innovate by introducing a synthetic element into this framework. Instead of solely using distorted versions of the same real image, we integrate a distorted version of a synthetic image. Following the notation from the previous section, let $z_i$ represent the distorted version of a real image and $\tilde{z_j}$ that of a synthetic image. The objective function for this adapted version of Barlow Twins, denoted as $\mathcal{L}_{BT}$, is formulated as:
\[\mathcal{L}_{BT}\triangleq \Sigma_i (1-\mathcal{C}_{ii})^2 + \lambda\, \Sigma_i \Sigma_{j\ne i}\,C_{ij}^2\]
Here, $\lambda$ is a positive constant that balances the first and second terms in the loss function. The cross-correlation matrix, $\mathcal{C}$, is computed between the outputs of the two identical networks, one fed with real images and the other with synthetic images, as follows:
\[\mathcal{C}_{ij}\triangleq \frac{\Sigma_b z_{b,i}\, \tilde{z}_{b,j}}{\sqrt{\Sigma_b (z_{b,i})^2} \sqrt{\Sigma_b (\tilde{z}_{b,j})^2 }}\]
where $b$ indexes the batch samples, while $i$ and $j$ index the vector dimensions of the networks' outputs. This updated approach, which integrates synthetic images into the Barlow Twins framework, focuses on aligning the representations of real ($z_i$) and synthetic images ($\tilde{z}_j$) via the cross-correlation matrix. We give more details regarding mixing in joint-embedding SSL in the appendix~\ref{supp:jssl}.

\subsection{Mixing in Distillation SSL}

\paragraph{DINO + MixDiff:} 
In contrast to other SSL methods, DINO~\cite{dino} uses a multi-crop strategy to create multiple views at different scales, including two high-resolution global views ($x^g_1$, $x^g_2$) and multiple lower-resolution local views ($x^l_k$, $k=1$ to $8$). It employs a knowledge distillation (KD) framework where a student network $g_{\theta_s}$ learns to match the output of a teacher network $g_{\theta_t}$, with the student processing both local and global views, while the teacher focuses on global views to enhance `local-to-global' learning. Building on this foundation, we introduce image mixing in DINO, termed DINO + MixDiff, the model is adapted to integrate both real and synthetic images. This is accomplished by adjusting the view composition to include one global and six local views from a real image, plus one global and two local views from a synthetic image. Consequently, our modified set includes a global view from a real image ($x^g_1$), a global view from a synthetic image ($\tilde{x}^g_2$), and four local views each from the real ($x^l_ r \;\forall\; r \in [1,6]$) and synthetic ($\tilde{x}^l_q \;\forall\; q \in [1,2]$) images.

With a student network $g(\theta_s)$ and a fixed teacher network $g{\theta_t}$ updated via \textit{Exponential Moving Average} (EMA), the learning objective is to align these distributions. This is achieved by minimizing the cross-entropy loss with respect to the student network's parameters $\theta_s$, expressed as: $\min_{\theta_s} H(P_t(X_t) , P_s(X_s))$
where $H(a,b) = -a\log b$ denotes the cross-entropy function. Both the student and teacher networks generate probability distributions denoted as $P_s$ and $P_t$, respectively,  derived by normalizing the networks' outputs using a softmax function. This learning process involves passing all crops through the student network, while the teacher network processes only the global views. This design fosters `local-to-global' as well as `real-to-synthetic' learning correspondences. The loss objective becomes: 
\[\min_{\theta_s}\;\; \Sigma_{X_t\in\{x^g_1, \tilde{x}^g_2\}} \;\Sigma_{X_s \in \tilde{V}, \; X_s\ne X_t}   H(P_t(X_t) , P_s(X_s))\]
We provide more details regarding mixing in distillation SSL such as DINO in appendix~\ref{supp:dino}.


\section{Experiments}
\label{sec:exp}

In this section, we present experiments testing robustness to distribution shifts, domain transfer across datasets, and performance on low-quality images. We provide insights into the learned representations using and without using MixDiff in SSL as described in Section~\ref{sec:method}.

\begin{table*}[t]
\begin{center}
\begin{small}
\begin{sc}
\begin{tabular*}{\linewidth}{@{\extracolsep{\fill}} lcccccccccc}
\toprule
\multirow{2}{*}{Model} & \multirow{2}{*}{IN-1K} & \multicolumn{7}{c}{Distribution Shifts Datasets} & \multirow{2}{*}{Mean} \\
\cmidrule{3-9}
 & & IN-V2 & IN-Sketch & IN-A & IN-R & ObjectNet & VW-C & Mean & \\
\midrule
SimCLR & 63.34 & 50.10 & 14.08 & 1.64 & 23.71 & 14.64 & 24.96 & 21.52 & 27.92 \\
SimCLR+Syn & 57.58 & 44.70 & 16.19 & 1.72 & 26.12 & 11.35 & 23.25 & 20.55 & 26.28 \\
SimCLR+MixDiff & \textbf{67.90} & \textbf{54.53} & \textbf{22.57} & \textbf{2.22} & \textbf{34.97} & \textbf{19.65} & \textbf{29.93} & \textbf{27.31} & \textbf{33.64} \\
\bottomrule
\end{tabular*}
\end{sc}
\end{small}
\end{center}
\vspace{-0.5cm}
\caption{Top-1 classification accuracies (\%) of models trained on ImageNet-1K, evaluated on domain shift datasets. SimCLR+Syn improves accuracy on IN-R and IN-Sketch, and this robustness extends to SimCLR+MixDiff. MixDiff enhances both in-distribution accuracy and out-of-distribution robustness.}
\label{robust-table-1k}
\end{table*}

\begin{table*}[t]
\begin{center}
\begin{small}
\begin{sc}
\begin{tabular*}{\linewidth}{@{\extracolsep{\fill} }lccccccccc}
\toprule
Model & CIFAR-10 & CIFAR-100 & Aircraft  & DTD & Flowers & Food & Pets37 & STL10 & Mean \\
\toprule
SimCLR & 79.75 & 55.49 & 30.48 & \textbf{62.71} & 81.27 & 64.80 & 67.51 & 93.75 & 66.97 \\

SimCLR+Syn  & 77.67 & 53.74 & 29.82 & 58.83 & 82.16 & 60.91 & 67.92 & 92.02 & 65.38 \\
SimCLR+MixDiff  & \textbf{84.76} & \textbf{64.33} & \textbf{36.84} & \textbf{62.71} & \textbf{88.43} & \textbf{66.40} &
\textbf{76.45} &
\textbf{95.80} &
\textbf{71.90}  \\

\bottomrule
\end{tabular*}
\end{sc}
\end{small}
\end{center}
\vspace{-0.5cm}
\caption{Comparison of transfer learning performance on eight diverse datasets for models trained on ImageNet-1K. MixDiff outperforms all other models, demonstrating superior generalization across these datasets}
\label{transfer-table}
\vspace{-0.4cm}
\end{table*}

\vspace{-0.35cm}
\paragraph{Training Algorithms and Data.} As the proposed solution is a simple change in one data branch of the SSL pipeline, we can easily incorporate it into any existing joint-embedding SSL methods. The substantial size of the ImageNet-1K (IN-1K) \cite{Deng2009ImageNetAL} dataset, which contains approximately 1.3 million images, presents challenges for extensive experimentation. Consequently, we primarily utilize the more manageable ImageNet-100 (IN-100) dataset \cite{Tian2019ContrastiveMC} for our studies, which include 100 classes and 1300 images per class. This dataset's smaller scale enables us to efficiently run multiple variations of each synthetic dataset and thoroughly evaluate the impact of various design choices. Nonetheless, we extend our experiments to IN-1K with SimCLR to validate our findings on a larger scale.

We trained variants of DINO, SimCLR, and Barlow Twins models using only real, synthetic, and our proposed mixed version of both image types (MixDiff) on the IN-100 dataset. For classification using the pre-trained features, unless stated, we always train the linear probes on the training set of real images. For example, the models trained with synthetic IN-100 images use the training set from real IN-100 images to train the linear probes to evaluate them on the IN-100 test set and other datasets. We leverage FFCV~\cite{leclerc2023ffcv} to accelerate training. We provide further details regarding configurations in appendix~\ref{supp:conf}.

\subsection{MixDiff boosts robustness to distribution shifts}
\begin{figure}
     \centering
     \includegraphics[width=\linewidth]{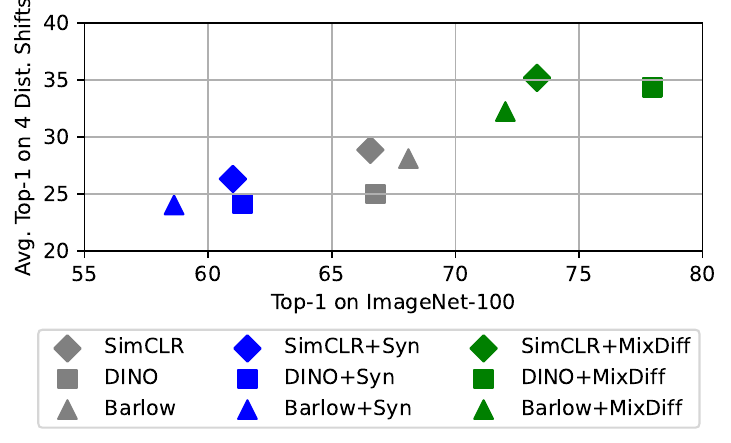}
     \vspace{-0.7cm}
     \caption{Top-1 classification accuracies (\%) for various models on ImageNet-100 (x-axis) and the average of four domain shift datasets (y-axis). This figure compares the performance of models trained on real, synthetic (Syn), and an equal combination of real and synthetic images (MixDiff). Models in the top-right quadrant exhibit better in-distribution and out-of-distribution accuracies.}
     \label{fig:robust}
    \vspace{-0.4cm}
     
\end{figure}

To evaluate performance under domain shifts, we choose a set of four datasets including ImageNet-A (IN-A) \cite{Hendrycks2019NaturalAE}, ImageNet-Sketch (IN-Sketch) \cite{Wang2019LearningRG}, ObjectNet \cite{Barbu2019ObjectNetAL}, ImageNet-V2 (IN-V2) \cite{Recht2019DoIC}, VizWiz-Classification (VW-C)~\cite{AkbarianBafghi2023AND}, and ImageNet-R (IN-R) \cite{Hendrycks2020TheMF}. 

 
Figure~\ref{fig:robust} shows the average accuracy on the four distribution shift datasets (excluding ObjectNet and WV-C due to the lack of common objects with IN-100) and compares it to the in-distribution IN-100. Models utilizing MixDiff (green) demonstrate superior accuracy on both IN-100 and the distribution shift datasets on average, outperforming models trained exclusively on real (grey) or synthetic (blue) images. This suggests that MixDiff not only enhances in-distribution performance but also enhances robustness against distribution shifts. SimCLR+MixDiff is the most robust, while DINO+MixDiff excels in in-distribution accuracy.
In Table \ref{robust-table-1k}, we drill down on the SimCLR variants using IN-1K dataset, and these findings align with our previous observations from the IN-100 dataset. While DINO outperforms on IN-100, we chose SimCLR for this experiment owing to its markedly quicker training time, attributed to its use of fewer crops and simpler overall setup. When pre-trained on ImageNet-1K, we observe similar improvements using mixing in SimCLR (SimCLR+MixDiff). Additionally, The performance boost is consistent across the four distribution-shift datasets. 

SimCLR trained on synthetic images (SimCLR+Syn) shows greater robustness on datasets like IN-Sketch and IN-R compared to SimCLR trained on real images. This improvement can be attributed to the datasets exhibiting properties similar to synthetic images. For instance, IN-Sketch contains human-drawn sketches, and IN-R includes artistic and stylized object renditions. These characteristics align well with the diverse and sometimes abstracted nature of the synthetic images. We believe MixDiff’s effectiveness is due to synthetic images acting as hard positive samples. It is more challenging to bring generated images closer than to bring augmented samples closer. These hard positive samples prevent the model from learning trivial features, which enhances its ability to learn effective representations~\cite{robinson2021can, wu2023synthetic}. However, it is crucial to highlight that accuracy on more challenging datasets like ImageNet-A is still low \cite{tomasev2022pushing,Djolonga2020OnRA}. This may be attributed to the backbone of the SimCLR model being ResNet-50. While ImageNet-A was curated specifically as images that fool the ResNet-50 model \cite{
Hendrycks2019NaturalAE}. More information and detailed numerical results are available in the appendix~\ref{robu-supp}.
\subsection{MixDiff improves transfer learning}
We evaluate the feature generality of the models by conducting transfer learning experiments across various image datasets and compare MixDiff's effectiveness with other models. The datasets include: Aircraft \cite{Maji2013FineGrainedVC}, DTD~\cite{Cimpoi2013DescribingTI}, Flowers102 \cite{Nilsback2008AutomatedFC}, Food101~\cite{Bossard2014Food101M}, Pets37~\cite{Parkhi2012CatsAD}, STL10~\cite{Coates2011AnAO}, CIFAR-10 \cite{Krizhevsky2009LearningML}, and CIFAR-100 \cite{Krizhevsky2009LearningML}.  

We pre-train the SimCLR model on the IN-1K dataset using the three data variants: real images, synthetic images, and MixDiff. For each variant, after pre-training the SimCLR backbone, we subsequently froze these layers to train linear probes on real data from the domains listed above. Table \ref{transfer-table} presents the top-1 accuracy results for each dataset. Our approach consistently achieves higher top-1 accuracy across all datasets compared to models trained solely on real or synthetic images. We would like to point out that our training iterations are lower (100 epochs) than the original SimCLR (1000 epochs), resulting in slightly lower numbers, but our method still consistently outperforms the original SimCLR model in relative terms. The superior transferability of representations learned using MixDiff can be attributed to its exposure to a broader range of visual inputs. By combining real and synthetic images, MixDiff allows the model to learn from a more diverse set of visual features and variations. This expanded visual vocabulary likely contributes to the development of more robust and generalizable representations, which in turn transfer more effectively to other datasets.

\begin{figure}
     \centering
     \includegraphics[width=\linewidth]{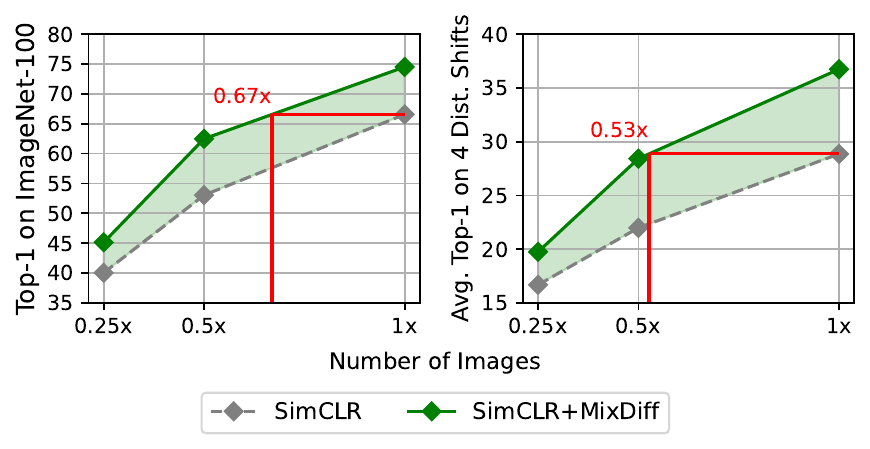}
     \vspace{-0.7cm}
     \caption{\textbf{Left:} Top-1 accuracy on IN-100 for SimCLR models trained with and without MixDiff at different scales of training images. \textbf{Right:} Average top-1 accuracy on four distribution shift datasets. SimCLR+MixDiff outperforms SimCLR, as indicated by the green area showing the performance gap.}
     \label{fig:limit}
          \vspace{-0.4cm}
\end{figure}

\subsection{MixDiff learns more from limited data}
We explore MixDiff's effectiveness in limited data scenarios for SSL. We trained SimCLR models using both real images and MixDiff on 25\%, 50\%, and 100\% subsets of the IN-100 dataset, supplemented with proportional synthetic images generated at a guidance scale of 8. Linear probes were subsequently trained on the real images of each subset. As shown in Figure~\ref{fig:limit}, we observe noticeable performance gains across different data-size regimes. As an example, using 25\% of the images, MixDiff achieves a 5.14\% increase in accuracy. This trend extends well to robustness, where a model trained with MixDiff on 50\% of the data matches the accuracy on distribution shift datasets of a model trained on 100\% real images.

Notably, as indicated by the red line in Figure~\ref{fig:limit}, MixDiff achieves comparable in-distribution accuracy and robustness to SimCLR trained on 100\% real images, while using only 53\% and 67\% of the real data, respectively. These findings demonstrate that MixDiff not only enhances performance but also enables robust training with reduced data requirements. It is important to note that while MixDiff offers significant efficiency gains in terms of data usage, it does require a one-time computational investment for generating synthetic images. We provide more discussion on the strategies for reducing training times using MixDiff with minimal performance impact in section~\ref{sec:compute} in the appendix.
\begin{figure}
     \centering
     \includegraphics[width=\linewidth]{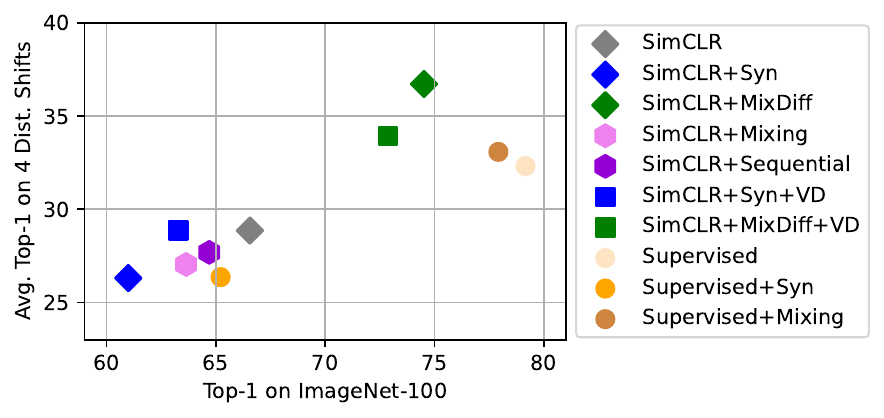}
     \vspace{-0.7cm}
     \caption{Top-1 classification accuracies for various models on ImageNet-100 (x-axis) and the average of four domain shift datasets (y-axis). The models use different approaches for mixing real and synthetic images, as well as varying generation models.}
     \label{fig:diffconfig}
        \vspace{-0.4cm}

\end{figure}
\vspace{-0.3cm}
\section{Ablation study}
Here, we empirically study the properties of generated synthetic images under varying guidance scales, mixing configurations, augmentations and evaluate how these factors affect on MixDiff's performance compared to other SSL methods. We further visualize self-attention maps of DINO over IN-100, and observe that DINO trained with MixDiff tend to attend and segment objects well with much less focus on the background (See section~\ref{sec:dino-mixing}). 

\subsection{Analysis of different configurations vs. MixDiff}
\label{abl:config}
We consider two new setups: training SimCLR on real images for 50 epochs followed by synthetic images for 50 epochs (Sequential), and training on a combined dataset of real and synthetic images for 50 epochs (Mixing), keeping the total number of images the same. Figure~\ref{fig:diffconfig} presents the performance of these new setups alongside our baseline configurations: SimCLR trained on synthetic images, real images, and our proposed MixDiff pipeline. While the new configurations surpassed SimCLR+Syn, they underperformed compared to SimCLR and SimCLR+MixDiff. We also replicated the MixDiff experiment using images generated by Versatile Diffusion (VD)\cite{Xu2022VersatileDT} with a guidance scale of 8, as an alternative to Stable Diffusion (SD)\cite{ivd}. Results in Figure~\ref{fig:diffconfig} demonstrate our MixDiff with VD-generated images (SimCLR+MixDiff+VD) outperforms the original SimCLR, indicating its generalizability across a different generative model. Finally, Figure~\ref{fig:diffconfig} shows the accuracy of a supervised ResNet-50 model trained on real, synthetic, and mixed data for 100 epochs, similar to the training setup of prior works~\cite{Azizi2023SyntheticDF,Yu2023DiversifyDF}. The model with real and mixed data achieves higher in-distribution accuracy, but SSL models trained with MixDiff are more robust to distribution shifts, which can be attributed to stronger SSL augmentations and learning from synthetic images. We show that data mixing can make the supervised model more robust at the cost of in-distribution accuracy, while MixDiff improves both. MixDiff narrows the gap between SimCLR and the supervised setting. Details on numerical results are in section~\ref{appendix-baselines} of the appendix.

\subsection{Impact of varying guidance scales}
\begin{figure}
      \centering
      \includegraphics[width=\linewidth]{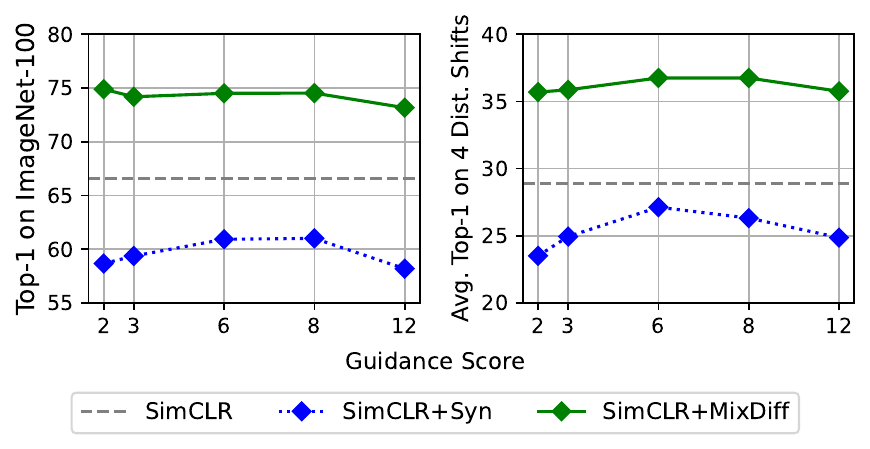}
      \vspace{-0.7cm}
      \caption{\textbf{Left:} Top-1 accuracy on IN-100. \textbf{Right:} Average top-1 accuracy on four distribution shift datasets. SimCLR+MixDiff is more robust to changes in guidance scale than SimCLR+Syn.}
      \label{fig:diffg}
           \vspace{-0.4cm}
\end{figure}

The guidance scale in generative models plays a crucial role in balancing the diversity and quality of synthesized images, which subsequently affects the learned representations, as demonstrated by~\cite{Tian2023StableRepSI}. To investigate this phenomenon in our context, we generated images using various guidance scales $\{2, 3, 6, 8, 12\}$ and trained different SimCLR configurations with these images. 

Figure~\ref{fig:diffg} illustrates our findings. Consistent with results from~\cite{Tian2023StableRepSI}, we observe that altering guidance scales impacts the robustness and in-distribution accuracy of models trained on synthetic images. This consistency is noteworthy, given our use of different generative models. Interestingly, MixDiff exhibits less variability to different guidance scales, maintaining a consistent performance across most scales, with a slight drop at a guidance scale of 12. This consistency offers a significant computational advantage, as it eliminates the need to fine-tune the guidance scale as a hyperparameter. Furthermore, our results consistently demonstrate that the SimCLR+Syn model underperforms compared to the original SimCLR. This finding indicates the constraints of using only synthetic data for training and suggests the advantages of our combined method. 
These findings collectively emphasize the importance of carefully considering guidance scale selection in generative models and suggest that our MixDiff method offers a more stable and efficient alternative for using synthetic data in representation learning tasks.

\subsection{Properties of generated images}
\begin{figure}
     \centering
     \includegraphics[width=\linewidth]{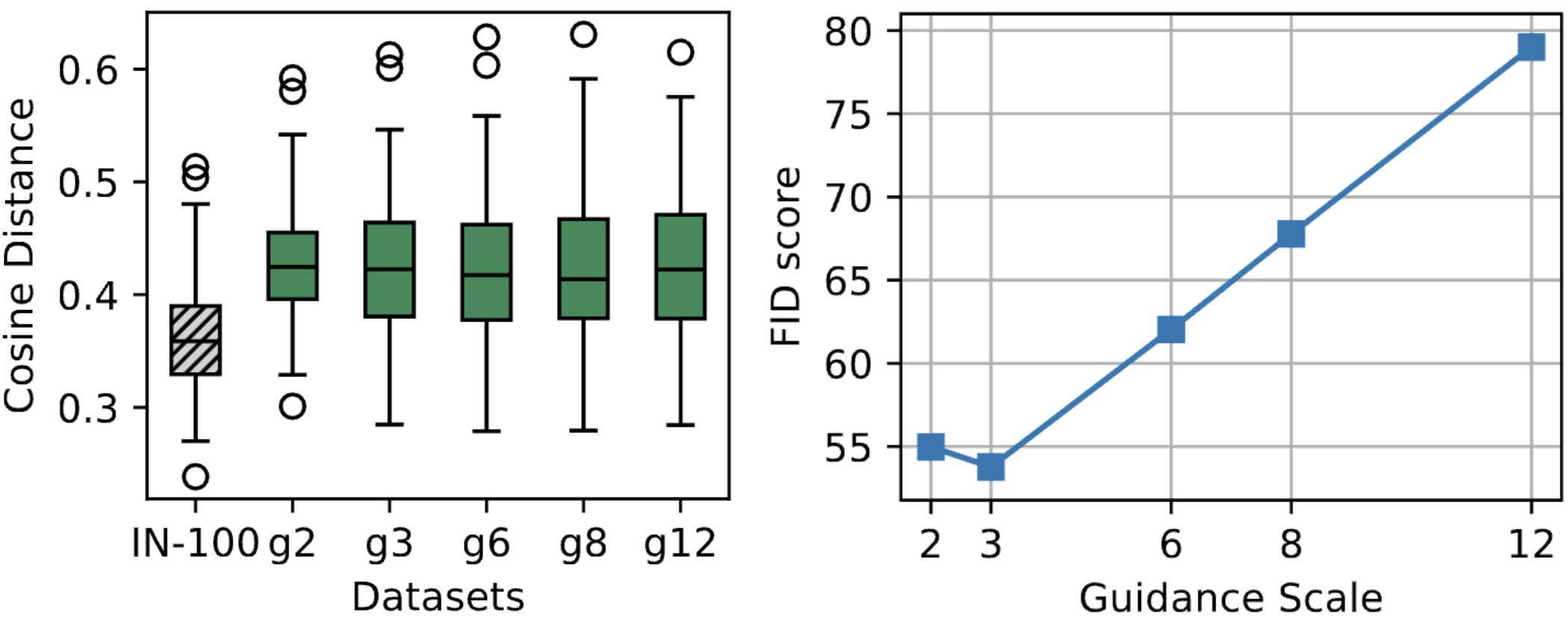}
     \vspace{-0.5cm}
     \caption{\textbf{Left:} Mean cosine distance distribution between feature vectors of IN-100 categories and generated images with varying guidance scales. \textbf{Right:} Relationship between guidance scales and FID scores. Generated images are more diverse than real images, and higher guidance scales reduce image quality.}
     \label{fig:fid}
     \vspace{-0.4cm}
\end{figure}

Building on our previous analysis of guidance effects on MixDiff's performance, we now investigate the properties of the generated synthetic images. We use cosine distance to measure diversity and Frechet Inception Distance (FID) scores \cite{Heusel2017GANsTB} to assess image quality in comparison to IN-100 across different guidance scales. The left panel of Figure~\ref{fig:fid} reveals that generated images exhibit higher diversity than real images, likely due to the generative model's inherent randomness. The right panel illustrates the similarity in quality between generated images and IN-100 source images across guidance scales. Drawing from the observations in previous section, we observe an inverse relationship between the SimCLR+Syn model's accuracy and image quality (as indicated by FID). For instance, a guidance scale of 8 yielded better performance despite a higher FID compared to a scale of 3 with the lowest FID. This aligns with findings from \cite{Tian2023StableRepSI}, suggesting that lower-quality generated images with higher FID scores may be more beneficial for learning solely from synthetic data. Interestingly, MixDiff's performance showed no clear trend relative to the FID scores of synthetic data. This robustness across guidance scales suggests that MixDiff may be effectively combining information from both real and synthetic sources, potentially offsetting image quality variations. Further research is needed to understand this phenomenon and its implications for using synthetic data in representation learning. 

\subsection{Impact of data augmentation}
Our study shows the effect of data augmentation on the SimCLR model, comparing its performance when trained with MixDiff versus solely real images of IN-100. Specifically, for models trained without augmentation, we retain only the random flip and remove all other augmentations.

\begin{figure}
     \centering
     \includegraphics[width=\linewidth]{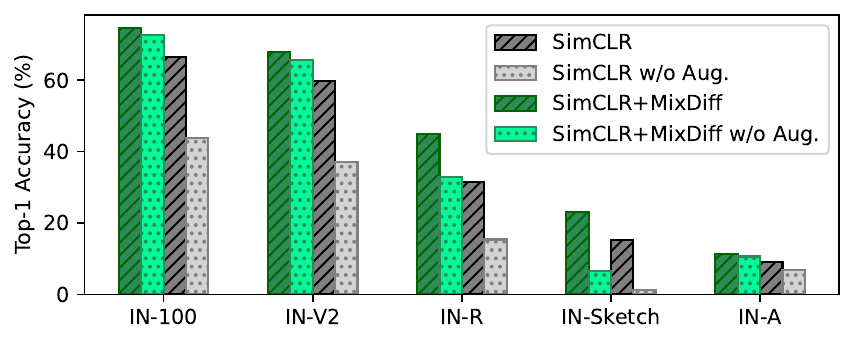}
     \vspace{-0.7cm}
     \caption{Top-1 accuracy comparison of SimCLR models trained with and without MixDiff, and the impact of including versus omitting SSL data augmentations.}
     \label{fig:aug}
     \vspace{-4mm}
\end{figure}

As depicted in Figure \ref{fig:aug}, an interesting finding is that omitting data augmentations, which is a key component in self-supervised learning models, does not significantly affect the performance of models trained with both real and synthetic images. Surprisingly, MixDiff model without augmentation generally outperforms the original SimCLR, on all except for the ImageNet-Sketch dataset. Conversely, SimCLR models trained exclusively on real images experience a significant drop in performance due to the absence of data augmentations. Specifically, the average accuracy drop across five validation datasets is 15.52\% for SimCLR without augmentation, compared to a lesser accuracy drop of 6.60\% for SimCLR+MixDiff. Notably, removing these augmentations can lead to faster training times, which is significant considering that data augmentations are often a major bottleneck in the training process, as highlighted in \cite{Bordes2023TowardsDJ}. Also, we observed that, for SimCLR+MixDiff, jittering is the most effective augmentation for in-distribution and robustness performance (See section~\ref{supp:augmentations}).
 
\section{Conclusion}
This work proposes MixDiff, a framework that demonstrates the potential of mixing synthetic images generated without any labels, with real images for fully self-supervised pre-training. We show that MixDiff consistently enhances the performance of existing SSL techniques across various benchmarks. Notably, MixDiff exhibits reduced variability to synthetic image quality and requires a smaller quantity of real data to achieve performance comparable to models trained on real data, thus offering a more robust and efficient SSL pre-training mechanism. More importantly, these results suggest that the integration of synthetic data with real images may serve as a viable alternative to augmentation techniques in existing SSL methods.

Further investigation reveals a noticeable performance gap between models trained solely on synthetic images and those trained on real images. This indicates a significant unexplored potential in the development of generative models to improve efficiency and produce images that more closely mimic the distribution of real images. In future work, we aim to investigate advanced generative diffusion models that could facilitate better mixing and enhance the robustness of visual features.

\paragraph{Acknowledgements.} This work utilized the Alpine high performance computing resource at the University of Colorado Boulder. Alpine is jointly funded by the University of Colorado Boulder, the University of Colorado Anschutz, Colorado State University, and the National Science Foundation (Award 2201538).

{\small
\bibliographystyle{ieee_fullname}
\bibliography{main}
}

\clearpage
\newpage
\appendix
\section{Overview}
This supplementary document enhances the primary paper in the following ways:
\begin{enumerate}
\item Provides additional insights and backgrounds into the methodology (complements \textbf{Section 3}).
\item Offers further details on training configurations, model setups, and the image generation procedure (complements \textbf{Section 4}).
\item Presents additional numerical results for robustness experiments (complements \textbf{Section 4.1}).
\item Provides further details on mixing in DINO and self-attention (complements \textbf{Section 5}).
\item Offers more information on computation costs (complements \textbf{Section 5}).
\item Presents additional numerical results for different configuration experiments (complements \textbf{Section 5.1}).
\end{enumerate}

\section{Background: Joint-embedding SSL}\label{supp:jssl}
One of the most successful self-supervised learning paradigms is joint-embedding SSL. Joint-Embedding Self-Supervised Learning (SSL) operates on the concept of instance-based views, where each image is treated as a distinct class~\cite{dosovitskiy2014discriminative, assran2023self}. This methodology capitalizes on data augmentation to generate diverse views of the same image. A notable subset of this approach is contrastive SSL techniques\cite{oord2018representation, moco, simclr}, which, within joint-embedding frameworks, aim to closely align the output embeddings of an image with those of its augmented version. Concurrently, these techniques strive to differentiate these embeddings from those of other images and their respective augmentations. Such methods are commonly implemented in Siamese network architectures~\cite{
chen2021exploring}, characterized by two parallel branches that are identical in structure and share the same weights. In our research, we focus on two joint-embedding frameworks: SimCLR\cite{simclr} and BarlowTwins\cite{barlow}. 

\subsection{Training and evaluating Joint-embedding SSL}
Given an unlabelled dataset $D = \{x_1, x_2...x_n\}$ where $x_i \in X \subset R^d$, represents the $i^th$ input. Any SSL method involves designing a pretext-task $S$ which utilizes pseudo-labels/feature representations $v_i \in Y$. We train a model $f(x)$ on $D_{train} \subset D$ such that $P(v|x)$, i.e. the conditional distribution of pseudo-labels given the input satisfies the pre-text task $S$. Eventually, we freeze the learned model parameters and train linear classifiers $g(f(x))$ on top of the model. The linear classifier is trained in a supervised manner via empirical risk minimization, $L(g(f(x)), y)$, where $y$ is the ground truth labels and $L$ is the training objective.
\begin{figure}
     \centering
     \includegraphics[width=\linewidth]{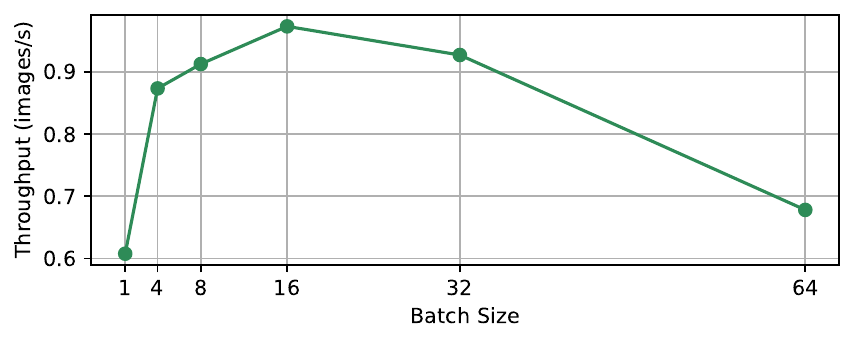}
     \vspace{-0.6cm}
     \caption{The throughput of the Stable Diffusion model used for image generation is optimized with a batch size of 16, which provides the best speed.}
     \vspace{-0.3cm}
     \label{fig:speed}
\end{figure}

\section{Method}
\subsection{Mixing in Distillation SSL}
Recent methods such as DINO~\cite{dino}, exploit architectural tricks inspired by knowledge distillation (KD)~\cite{hinton2015distilling}. The KD technique involves training a `student' network to emulate the output representations of a `teacher' network. The teacher network's weights are dynamically updated as a running average of the student network’s weights or are shared with the student network~\cite{byol}. Notably, in this configuration, gradients are not backpropagated through the teacher network~\cite{chen2021exploring}.

\begin{table*}[t]
\begin{center}
\begin{small}
\begin{sc}
\begin{tabular*}{\linewidth}{@{\extracolsep{\fill}} lccccccc}
\toprule
\multirow{2}{*}{Model} & \multirow{2}{*}{IN-100} & \multicolumn{5}{c}{Distribution Shifts Datasets} & \multirow{2}{*}{Mean} \\
\cmidrule{3-7}
 & & IN-V2 & IN-Sketch & IN-A & IN-R & Mean & \\
\midrule
SimCLR & 66.56 & 59.80 & 15.12 & 9.00 & 31.52 & 28.86 & 36.40 \\
SimCLR+Syn & 61.00 & 52.50 & 14.55 & 8.43 & 29.77 & 26.31 & 33.25 \\
SimCLR+MixDiff & \textbf{73.30} & \textbf{66.30} & \textbf{21.27} & \textbf{11.13} & \textbf{44.02} & \textbf{35.18} & \textbf{43.20} \\
\midrule
DINO & 66.76 & 57.90 & 10.13 & 8.52 & 27.46 & 25.00 & 34.15 \\
DINO+Syn & 61.38 & 54.30 & 8.68 & 7.49 & 27.97 & 24.11 & 31.96 \\
DINO+MixDiff & \textbf{77.96} & \textbf{70.20} & \textbf{18.89} & \textbf{12.85} & \textbf{37.50} & \textbf{34.36} & \textbf{43.48} \\
\midrule
Barlow & 68.10 & 62.90 & 12.25 & 8.12 & 31.14 & 28.10 & 36.50 \\
Barlow+Syn & 58.62 & 53.30 & 9.49 & 7.02 & 26.30 & 24.03 & 30.95 \\
Barlow+MixDiff & \textbf{72.02} & \textbf{64.40} & \textbf{16.38} & \textbf{9.42} &\textbf{38.71} & \textbf{32.23} & \textbf{40.19} \\
\bottomrule
\end{tabular*}
\end{sc}
\end{small}
\end{center}
\vspace{-0.5cm}
\caption{Top-1 classification accuracies (\%) for various models on ImageNet-100 domain shift datasets. This table compares the performance of models trained on real, synthetic (Syn), and an equal combination of real and synthetic (MixDiff) images.}
\label{tab:robust-table}
\end{table*}

\begin{table}[t]
\begin{center}
\begin{small}
\begin{sc}
\begin{tabular*}{\linewidth}{@{\extracolsep{\fill} }lccc}
\toprule
Model & Corrupted & Clean & Mean \\
\toprule
SimCLR&  18.87 & 30.16 & 24.96 \\
SimCLR+Syn&  18.79 & 28.11 & 23.25 \\
SimCLR+MixDiff&  \textbf{23.94} & \textbf{35.02} & \textbf{29.93} \\
\bottomrule
\end{tabular*}
\end{sc}
\end{small}
\end{center}
\vspace{-0.5cm}
\caption{Accuracies of models trained on ImageNet-1K, evaluated on corrupted, clean, and all images of VW-C.}
\label{quality-1k}
\end{table}
\vspace{-0.4cm}
\paragraph{DINO and DINO + MixDiff:}\label{supp:dino} 
The original implementation of DINO~\cite{dino} employs a multi-crop strategy to process multiple scaled views of an image. Specifically, for a given image, a set $V$ of different views is generated, comprising two global views ($x^g_1$ and $x^g_2$) at higher resolutions and multiple local views ($x^l_ k$ for $k$ ranging from 1 to 8) at lower resolutions. Consider he training framework for DINO leverages the following KD framework: a student network $g_{\theta_s}$, parameterized by $\theta_s$, is trained to align with the output of a teacher network $g_{\theta_t}$, parameterized by $\theta_t$. All views (local + global) are processed by the student, while only the global views are handled by the teacher, promoting a 'local-to-global' learning correspondence.

Building on this foundation, we introduce image mixing in DINO, termed DINO + MixDiff. In this modified approach, instead of using two global views and eight local views from each real image, we employ a mix of real and synthetic images. Specifically, we use one global view and six local views from a real image, supplemented by one global view and two local views from a corresponding synthetic image. This particular mix of real and synthetic views was determined empirically, and we explore varying levels of mixing and its impact on DINO model performance in Section~\ref{ablation:mixing}. Consequently, our modified set includes a global view from a real image ($x^g_1$), a global view from a synthetic image ($\tilde{x}^g_2$), and four local views each from the real ($x^l_ k \forall k \in [1,4]$) and synthetic ($\tilde{x}^l_ k \forall k \in [1,4]$) images. In this framework, both the student and teacher networks generate probability distributions over K dimensions, denoted as $P_s$ and $P_t$, respectively. These probabilities are derived by normalizing the networks' outputs using a softmax function. Specifically,

\[ P_s(X)^{(i)} = \frac{\exp(g_{\theta_s} (X)^{(i)}/\tau_s)}{ \Sigma_{k=1}^{K}\exp(g_{\theta_s} (X)^{(k)}/\tau_s)} \]

The parameter $\tau_s$, a positive temperature value, modulates the sharpness of the student network's output distribution. The set $X_s$ for the student encompasses all views, namely $\{ x^g_1, \tilde{x}^g_2, x^{l}_k, \tilde{x}^{l}_k\} \forall k \in [1,4]$, named as a set $\tilde{V}$. In a parallel manner, the teacher network's output distribution is influenced by its own temperature parameter $\tau_t$. However, the set $X_t$ for the teacher is restricted to only global views $\{ x^{g}_1, \tilde{x}^g_2 \}$.

\section{Experiments}\label{supp:exp}
This section provides configurations and numerical results for the robustness of different models to distribution shifts.

\subsection{Configurations}
\label{supp:conf}
\paragraph{Training.} Our training employs FFCV-SSL \cite{Bordes2023TowardsDJ}, an extended version of the FFCV library \cite{leclerc2023ffcv} with added support for SSL data augmentations, enhancing training speed and feasibility within our limited resources. We have adapted the code for the DINO framework, in addition to Barlow Twins and SimCLR models. For these models, the LARS \cite{You2017LargeBT} optimizer with a learning rate of 1 is used, while DINO models utilize the AdamW \cite{Loshchilov2017DecoupledWD} optimizer with a learning rate of $0.0005$. All experiments are conducted with a consistent batch size of 256, employing the Cosine Decay scheduler \cite{Loshchilov2016SGDRSG}. Offline linear probes are conducted with the AdamW optimizer at a learning rate of 0.0001 across all experiments, utilizing only real images for training in every case.
\vspace{-0.4cm}
\paragraph{Fine-tuning.} For transfer learning, we tested learning rates of $\{0.5, 0.05, 0.01, 0.005\}$ and trained the models for 15,000 steps, assessing accuracy every 500 steps. The checkpoint with the highest accuracy was selected as the best model for each experiment. We fine-tuned only linear probes while keeping the base model frozen.
\vspace{-0.4cm}
\paragraph{Models.}  In our experiments, SimCLR and Barlow Twins utilize ResNet-50 as the base encoder, while DINO employs Vision Transformer Small/16 (ViT-S/16) as its backbone. Additionally, a linear layer mapping the representation size to the class count is added atop the encoder. The representation size for SimCLR and Barlow Twins is 2048, whereas for DINO, it's set to 65536, identified as optimal for DINO \cite{dino}.
\vspace{-0.4cm}
\paragraph{Generating images.} Our process for generating images utilizes 50 diffusion steps, as noted in \cite{Sariyildiz2022FakeIT}, expanding the step count beyond 50 appears to be unnecessary. For the IN-100 dataset, we use various guidance scales $\{2,3,6,8,12\}$ to generate images. Following the findings of \cite{Azizi2023SyntheticDF}, we produce large images with a resolution of $512 \times 512$ pixels, batched in groups of 16. As indicated in Figure~\ref{fig:speed}, we determined the optimal throughput for image generation by testing different batch sizes on an NVIDIA A100 GPU. We also deactivate the safety checker in the package to prevent the generation of blank images. Generating images for IN-100 takes approximately 35 hours, while for IN-1K, it takes around 365 hours using a single GPU. For the IN-1K dataset, a consistent guidance scale of 8 is applied. We set the random seed to 25 for reproducibility. We generate images before training. Additionally, for generating images with Versatile Diffusion (VD)~\cite{Xu2022VersatileDT}, we use a guidance scale of 8. 

\subsection{MixDiff boosts robustness to distribution shifts}
\label{robu-supp}
We present results from experiments on IN-100. Table~\ref{tab:robust-table} details the accuracy of SimCLR, DINO, and Barlow Twins models trained on real, synthetic, and blended images. We also evaluate pre-trained models on the IN-1K dataset using the VizWiz-Classification (VW-C) dataset \cite{AkbarianBafghi2023AND}, which contains images with various quality issues taken by blind individuals. We assess performance on corrupted (mean accuracy of six quality issues), clean, and all images. Table \ref{quality-1k} shows our results. Notably, the model trained on synthetic data shows a smaller drop in accuracy on corrupted images (0.08\%) compared to clean VW-C images (2.05\%) and the ImageNet validation dataset (6.06\%). Our framework enhances SimCLR models' ability to handle quality issues, demonstrating its practical value in real-world scenarios.
\begin{table*}[t]
\begin{center}
\begin{small}
\begin{sc}
\begin{tabular*}{\linewidth}{@{\extracolsep{\fill}} lccccccc}
\toprule
\multirow{2}{*}{Model} & \multirow{2}{*}{IN-100} & \multicolumn{5}{c}{Distribution Shifts Datasets} & \multirow{2}{*}{Mean} \\
\cmidrule{3-7}
 & & IN-V2 & IN-Sketch & IN-A & IN-R & Mean & \\
\midrule
Supervised & \textbf{79.16} & \textbf{71.10} & 11.43 & 10.20 & 36.55  & 32.32 & 41.69 \\
Supervised+Syn & 65.22&58.30&9.05&8.38&29.73	&26.36&34.14 \\
Supervised+Mixing & 77.92 & 69.40 & 13.96 & \textbf{10.98} & 37.97	&33.08&\textbf{42.05} \\
\midrule
SimCLR & 66.56 & 59.80 & 15.12 & 9.00 & 31.52 & 28.86 & 36.40 \\
SimCLR+Mixing & 63.64 & 56.30 & 14.51 & 8.22 & 29.13 & 27.04 & 34.36 \\
SimCLR+Sequential & 64.70&57.70&14.34&8.12&30.61&27.69&35.09\\
\midrule
SimCLR+Syn+VD & 63.28 & 54.50 & 18.15 & 9.37 & 33.51 & 28.88 & 33.25 \\
SimCLR+MixDiff+VD & 72.88 & 66.70 & \textbf{18.75} & 10.46 & \textbf{39.81} & \textbf{33.93} & 41.72 \\
\bottomrule
\end{tabular*}
\end{sc}
\end{small}
\end{center}
\vspace{-0.5cm}
\caption{Top-1 classification accuracies (\%) for various models on ImageNet-100 and domain shift datasets. This table compares the performance of models trained on real, synthetic images (Syn), and an equal combination of real and synthetic images (MixDiff). The synthetic images were generated using a different generative model, Versatile Diffusion (VD), with a guidance scale of 8. Also, we report the performance of supervised models trained on our synthetic data and real data.}
\label{tab:another-gen}
\end{table*}

\begin{table*}[t]
\begin{center}
\begin{small}
\begin{sc}
\begin{tabular*}{\linewidth}{@{\extracolsep{\fill}} llccccccc}
\toprule
\multirow{2}{*}{Model} & \multirow{2}{*}{Augmentation} & \multirow{2}{*}{IN-100} & \multicolumn{5}{c}{Distribution Shifts Datasets} & \multirow{2}{*}{Mean} \\
\cmidrule{4-8}
 & & & IN-V2 & IN-Sketch & IN-A & IN-R & Mean & \\
\midrule
\multirow{6}{*}{SimCLR+MixDiff} & None & 66.34 & 58.10 & 4.91 & 10.46 & 29.41 & 25.50 & 33.67 \\
& Flip & 67.08 & 60.00 & 4.48 & 10.20 & 29.47 & 26.04 & 34.25 \\
& Blur & 65.94 & 56.40 & 4.62 & \textbf{10.41} & 30.87 & 25.57 & 33.65 \\
& Solarization  & 65.18 & 57.60 & 5.77 & 9.31 & 30.30 & 25.75 & 33.63 \\
& Jitter & \textbf{67.92} & 60.30 & 7.46 & 9.16 & 33.13 & 27.51 & 33.59 \\
& All & 67.16 & \textbf{60.70} & \textbf{17.99} & 10.09 & \textbf{37.36} & \textbf{31.53} & \textbf{38.66} \\
\bottomrule
\end{tabular*}
\end{sc}
\end{small}
\end{center}
\vspace{-0.5cm}
\caption{Top-1 classification accuracies (\%) for various models on ImageNet-100 and domain shift datasets. This table compares the performance of SimCLR+MixDiff trained with different augmentations after 50 epochs. The results show that the first row, which does not have any augmentation, has comparable performance to the last row, which leverages all the augmentations.}
\label{tab:differentaug}
\vspace{-0.4cm}
\end{table*}

\begin{figure}
    \centering
    \includegraphics[width=\linewidth]{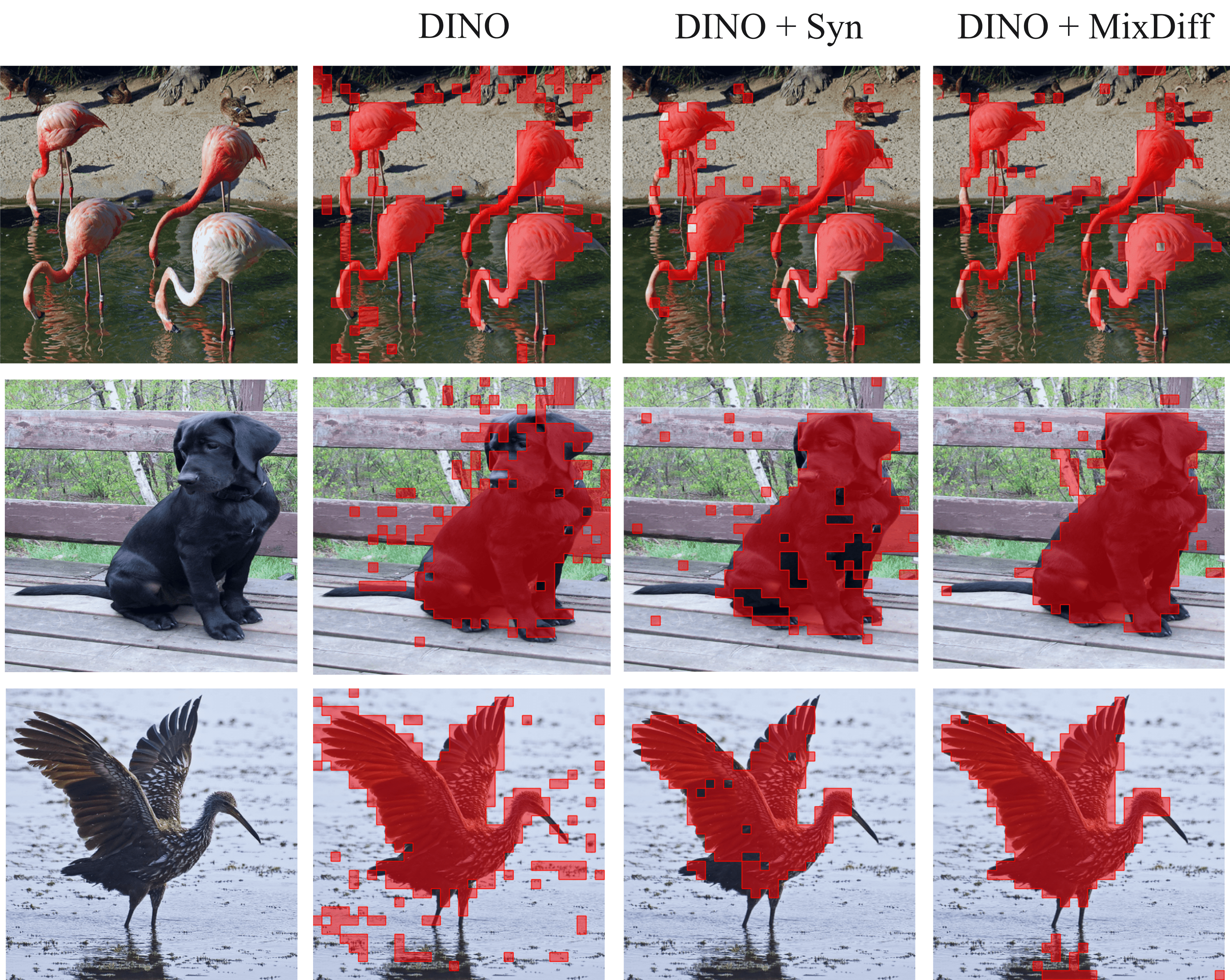}
    \vspace{-0.5cm}
    \caption{We present the masks generated by applying a 65\% threshold to the self-attention maps. The above maps were obtained using the best head from each ViT-S/16 DINO model that has been trained on ImageNet-100 using real and synthetically generated ImageNet versus MixDiff.}
    \label{fig:attn}
    \vspace{-0.5cm}
\end{figure}

\begin{table}[t]
\begin{center}
\begin{small}
\begin{sc}
\begin{tabular*}{\linewidth}{@{\extracolsep{\fill} }cccccc}
\toprule
\multirow{2}{*}{MixDiff} &
\multicolumn{2}{c}{Real} & 
\multicolumn{2}{c}{Syn} & 
\multirow{2}{*}{Acc.}  \\
\cline{2-3} \cline{4-5}  
&\#L & \#G & \#L & \#G & \\
\toprule

$\times$ & 8 & 2 & - & - & 66.75 \\
$\times$ & - & - & 8 & 2 & 61.38 \\
$\surd$ & 2 & 1 & 6 & 1 & 77.43 \\
$\surd$ & 4 & 1 & 4 & 1 & 77.52 \\
$\surd$ & 6 & 1 & 2 & 1 & \textbf{77.96} \\

\bottomrule
\end{tabular*}
\end{sc}
\end{small}
\end{center}
\vspace{-0.5cm}
\caption{Comparison of different configurations of the DINO model and their top-1 accuracy on the IN-100 validation dataset. The first row presents the original settings. Columns represent the number of local (\#L) and global (\#G) views.}
\label{dino-table}
\vspace{-0.4cm}
\end{table}

\section{Ablation Study}

\subsection{DINO: Impact of mixing in self-attention and performance}
\label{sec:dino-mixing}
In our analysis of DINO~\cite{dino} models trained on IN-100, we interpret masks derived from self-attention maps by applying thresholds for enhanced visualization. These maps, sourced from the top-performing head of each ViT-S/16~\cite{Dosovitskiy2020AnII} DINO model trained on both real and synthetic ImageNet datasets using our MixDiff approach, are not designed for mask creation but rather to highlight the model's focus areas during image processing. Our findings, illustrated in Figure~\ref{fig:attn}, show that models trained on real images effectively segment objects with some background attention. In contrast, the model trained on synthetic images shows a tendency to focus less on the background, but the overall object segmentation appears somewhat less defined. The MixDiff-trained model strikes a balance, demonstrating clearer object focus with minimal background distraction, indicating improved object segmentation. This improvement is clearly visible in the last row of Figure~\ref{fig:attn}. This suggests that integrating synthetic data with the MixDiff method may enhance scene understanding and image segmentation capabilities.

\subsection{Performance trend with varying mixing levels}\label{ablation:mixing}
We explore various configurations of image splitting between real and synthetic images for the DINO model. DINO utilizes a total of 10 crops: the teacher model processes two global crops, while the student model analyzes eight local views in addition to the global views. Table \ref{dino-table} presents a comparison of different configurations and their corresponding top-1 accuracy on the IN-100 validation dataset. All synthetic images are generated with a guidance scale of 8.

The findings demonstrate that dividing the training dataset into a mix of real and synthetic images can enhance accuracy, and our framework shows robustness to varying proportions of these images. The most effective configuration identified involves using 2 local views and 1 global view from synthetic images, along with 6 local views and 1 global view from real images. This specific arrangement yielded the highest top-1 accuracy, reaching 78.24\%.

\subsection{Impact of data augmentation}
\label{supp:augmentations}
This experiment examines various MixDiff configurations, including mixing, sequential, and versatile diffusion (definition borrowed from Section 5.1), comparing them against supervised baselines. We use a ResNet-50 architecture for our supervised baseline, training it with standard, synthetic, and mixing approaches. Our evaluation covers both the ImageNet-100 (IN-100) test set and several distribution shift datasets. Our findings reveal that the standard supervised ResNet-50 model performs best on the IN-100 test set but slightly underperforms on distribution shift datasets. Interestingly, the SimCLR model trained with MixDiff (versatile diffusion) achieves the highest average performance across distribution shift datasets, suggesting enhanced generalization capabilities. The supervised model trained with mixing shows the best performance on ImageNet-A (IN-A), possibly due to the dataset's synthetic characteristics. The SimCLR model with MixDiff (versatile diffusion) also demonstrates comparable performance on IN-A.


\subsection{Additional baselines}
\label{appendix-baselines}
We take variants of MixDiff like mixing, sequential and versatile diffusion borrowed from section~5.1 and compare them with supervised baselines. For supervised ResNet-50, we use train and evaluate all three variants (standard, synthetic and mixing). Table~\ref{tab:another-gen} summarizes these results. We observe that while standard supervised has the best performance on IN-100 test set, it lags a little behind in distribution shift datasets. SimCLR trained with MixDiff (versatile diffusion) outperforms both SimCLR and SimCLR+Syn, demonstrating the effectiveness of our method using different generative model.

\subsection{Computational Costs}
\label{sec:compute}

\begin{figure}
     \centering
     \includegraphics[width=0.8\linewidth]{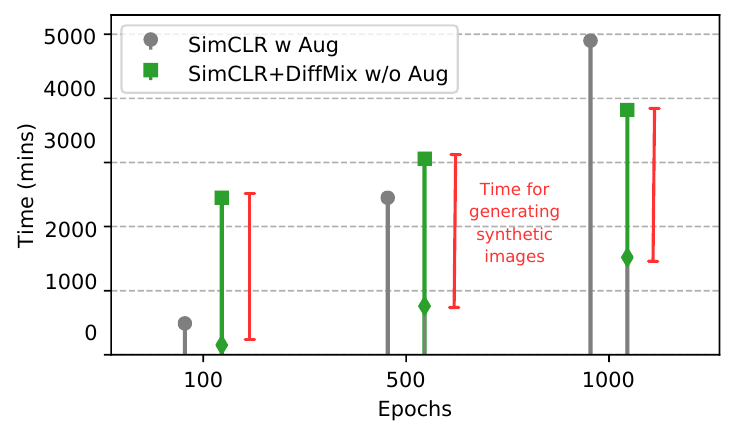}
     \vspace{-0.5cm}
     \caption{Total compute time trend comparison with increasing epochs for SimCLR, SimCLR+MixDiff with and w/o augmentations, respectively.}
     \label{fig:time}
     \vspace{-0.4cm}
\end{figure}

One limitation of using synthetic data is computation time. However, as discussed in Section~4.3, our method is beneficial in limited data scenarios and needs only one generated image per real image. Incorporating MixDiff into SSL pre-training initially slows it down due to synthetic image generation. We compare SimCLR's training time with augmentations to SimCLR+MixDiff without augmentations on an NVIDIA A100 GPU with a batch size of 256. Using MixDiff without additional augmentations results in only a minimal performance drop, suggesting that synthetic images can complement natural images without needing augmentations. This synthetic image generation process is a one-time upfront cost (amount of time shown in red), unlike traditional augmentations that incur computational costs at each training iteration. Once synthetic images are generated, training MixDiff without any augmentations is super efficient. This gap can be seen as the difference between the circular grey point and the lower diamond green point. Figure~\ref{fig:time} shows that MixDiff's one-time costs scale better compared to SimCLR, whose augmentations incur costs at each iteration.  We recommend using MixDiff with traditional augmentations for the best performance. The enhanced quality of learned representations and improved generalization across diverse scenarios justify the computational trade-off, especially where robustness and adaptability are crucial.

\end{document}